\documentclass{IOS-Book-Article}

\usepackage{mathptmx}
\usepackage{soul}
\PassOptionsToPackage{hyphens}{url}\usepackage{hyperref}
\usepackage{listings}
\usepackage{kg-macros}
\usepackage{cube}
\usepackage{graphicx}
\usepackage{times}
\usepackage{latexsym}
\usepackage{booktabs}
\usepackage[nolist]{acronym}
\usepackage{todonotes}
\usepackage{array}
\usepackage{mdwtab}
\usepackage{tabularx}
\usepackage{dcolumn}
\usepackage{longtable}
\usepackage{supertabular}
\usepackage[inline]{enumitem}
\usepackage{xspace}
\usepackage{fancyvrb}
\usepackage{comment}
\usepackage{wrapfig}  
\usepackage{caption}
\usepackage{longtable}
\usepackage{multirow}
\usepackage{colortbl}
\usepackage[T1]{fontenc}
\usepackage{xcolor}
\usepackage{pgfplots}
\usepackage{subcaption}
\usepackage{footnote}
\usepackage{amssymb}
\usepackage{mathtools}
\usepackage{verbatim}
\usepackage{lipsum}
\usepackage{listingsutf8}
\usepackage{lscape}
\usepackage{pifont}
\usepackage{numprint}
\usepackage{adjustbox}
\usepackage{multirow}
\usepackage{multicol}
\usepackage{float}
\usepackage{textcomp}
\usepackage{csvsimple}
\usepackage{ulem}
\def\hb{\hbox to 10.7 cm{}}
\newtheorem{theorem}{Theorem}
\begin{document}

\pagestyle{headings}
\def\thepage{}

\begin{frontmatter}              

\title{Knowledge Graph Question Answering using Graph-Pattern Isomorphism}

\markboth{}{September 2021 \hb}

\author[A]{\fnms{Daniel} \snm {Vollmers}}
\author[A]{\fnms{Rricha} \snm{Jalota}}
\author[A,D]{\fnms{Diego} \snm{Moussallem}}
\author[A]{\fnms{Hardik} \snm{Topiwala}}
\author[A]{\fnms{Axel{-}Cyrille} \snm{Ngonga Ngomo}}
and
\author[B,C]{\fnms{Ricardo} \snm{Usbeck}}

\runningauthor{Vollmers et al.}
\address[A]{Data Science Group, Paderborn University, Germany}
\address[B]{Fraunhofer IAIS, Dresden, Germany}
\address[C]{Universität Hamburg, Germany}
\address[D]{Globo, Rio de Janeiro, Brazil}

\begin{abstract}
Knowledge Graph Question Answering (KGQA) systems are often based on machine learning algorithms, requiring thousands of question-answer pairs as training examples or natural language processing pipelines that need module fine-tuning.
In this paper, we present a novel QA approach, dubbed TeBaQA. Our approach learns to answer questions based on graph isomorphisms from basic graph patterns of SPARQL queries. Learning basic graph patterns is efficient due to the small number of possible patterns.
This novel paradigm reduces the amount of training data necessary to achieve state-of-the-art performance. TeBaQA also speeds up the domain adaption process by transforming the QA system development task into a much smaller and easier data compilation task.
In our evaluation, TeBaQA achieves state-of-the-art performance on QALD-8 and delivers comparable results on QALD-9 and LC-QuAD v1.
Additionally, we performed a fine-grained evaluation on complex queries that deal with aggregation and superlative questions as well as an ablation study, highlighting future research challenges.
\end{abstract}

\end{frontmatter}
\markboth{September 2021\hb}{September 2021\hb}
\section{Introduction}
The goal of most Knowledge Graph (KG) Question Answering (QA) systems is to map natural language questions to corresponding SPARQL queries. 
This process is known as semantic parsing~\cite{berant2013semantic} and can be implemented in various ways. 
A common approach is to utilize query templates (alias graph patterns) with placeholders for relations and entities. 
The placeholders are then filled with entities and relations extracted from a given natural language question~\cite{DBLP:conf/www/AbujabalRYW18,DBLP:journals/semweb/HoffnerWMULN17,TBSL} to generate a SPARQL query, which is finally executed.  
Semantic parsing assumes that a template can be constructed or chosen to represent a natural language question's internal structure.
Thus, the KGQA task can be reduced to finding a matching template and filling it with entities and relations extracted from the question.

The performance of KGQA systems based on this approach depends heavily on the implemented query templates, which depend on the question's complexity and the KG's topology.
Consequently, costly hand-crafted templates designed for a particular KG cannot be easily adapted to a new domain. 

In this work, we present a novel KGQA engine, dubbed TeBaQA.
TeBaQA alleviates manual template generation's effort by implementing an approach that learns templates from existing KGQA benchmarks. We rely on learning of templates based on isomorphic basic graph patterns. 

The goal of TeBaQA is to employ machine learning and feature engineering to learn to classify natural language questions into isomorphic basic graph pattern classes.
At execution time, TeBaQA uses this classification to map a question to a basic graph pattern, i.e., a template, which it can fill and augment with semantic information to create the correct SPARQL query.

TeBaQA achieves state-of-the-art performance partially without any manual effort. 
In contrast to existing solutions, TeBaQA can be easily ported to a new domain using only benchmark datasets, partially proven by our evaluation over different KGs and train-test-sets. We use a best-effort to work with the data at hand instead of either (i) requiring a resource-intensive dataset creation and annotation process to train deep neural networks or (ii) to hand-craft mapping rules for a particular domain.
Our contributions can be summarized as follows:
\begin{itemize}
    \item We present TeBaQA, a QA engine that learns templates from benchmarks based on isomorphic basic graph patterns.
    \item We describe a greedy yet effective ranking approach for query templates, aiming to detect the best matching template for a given input query. 
    \item We evaluate TeBaQA on several standard KGQA benchmark datasets and unveil choke points and future research directions.
    
\end{itemize}

The code is available at \url{https://github.com/dice-group/TeBaQA} and a demo of TeBaQA over encyclopedic data can be found at \url{https://tebaqa.demos.dice-research.org/}.
We also provide an online appendix which contains more details about our algorithms and their evaluations at \url{https://github.com/dice-group/TeBaQA/blob/master/TeBaQA_appendix.pdf}.

\section{Related Work}
The domain of Knowledge Graph Question Answering has gained traction over the last few years. There has been a shift from simple rule-based systems to systems with more complex architectures that can answer questions with varying degrees of complexity. In this section, we provide an overview of the recent work. 
We begin with approaches that took part in QALD challenge series~\cite{qald-7,qald-8,qald-9}.

gAnswer2~\cite{DBLP:conf/sigmod/ZouHWYHZ14} addresses the translation of natural language to SPARQL as a subgraph matching problem. 
Following the rule-based paradigm, QAnswer~\cite{DBLP:journals/semweb/DiefenbachGBSM20} utilizes the semantics embedded in the underlying KG (DBpedia and Wikidata) and employs a combinatorial approach to create SPARQL queries from natural language questions. QAnswer overgenerates possible SPARQL queries and has to learn the correct SPARQL query ranking from large training datasets.

Second, we introduce approaches that rely on templates to answer natural questions over knowledge bases.
Hao et al.~\cite{DBLP:conf/coling/HaoLHLZ18} introduced a pattern-revising QA system for answering simple questions. It relies on pattern extraction and joint fact selection, enhanced by relation detection, for ranking the candidate subject-relation pairs. 
NEQA~\cite{DBLP:conf/www/AbujabalRYW18} is a template-based KBQA system like TeBaQA, which uses a continuous learning paradigm to answer questions from unseen domains. Apart from using a similarity-based approach for template-matching, it relies on user feedback to improve over time. QUINT~\cite{Abujabal:2017:ATG:3038912.3052583} also suggests a template learning system that can perform on compositional questions by learning sub-templates.
KBQA~\cite{DBLP:journals/pvldb/CuiXWSHW17} extracted 27 million templates for 2782 intents and their mappings to KG relations from a QA corpora (Yahoo! Answers). 
The way these templates are generated and employed differs from that of TeBaQA. NEQA relies on active learning on sub-parts and thereby, possibly misses the semantic connection between question parts. KBQA learns many templates and can hence fail to generalize well to other domains or KB structures. 

In 2019, Zheng et al.~\cite{zheng2019question} use structural query pattern, a coarse granular version of SPARQL basic graph patterns which they later augment with different query forms and thus can generate also SPARQL query modifiers.
The closest work to ours is by Athreya et al.~\cite{athreya2020templatebased} based on a tree-based RNN to learn different templates on LC-QuAD v1 which the authors directly derive from the LC-QuAD v1 inherent SPARQL templates and thus cannot generalize to other KGs or datasets. 
CONQUEST~\cite{DBLP:conf/nldb/AvilaFMV20} is an enterprise KGQA system which also assumes the SPARQL templates are given. It then matches the questions and templates by vectorizing both and training one classifier, namely Gaussian Naïve Bayes.

QAMP~\cite{DBLP:conf/cikm/VakulenkoGPRC19} is an unsupervised message passing system using a similarly simple approach to information extraction as TeBaQA. QAMP outperforms QAnswer on LC-QuAD v1.
Recently, Kapanipathi et al.~\cite{DBLP:journals/corr/abs-2012-01707} present a system, dubbed NSQA, which is in pre-print. NSQA is based on a combination of semantic parsing and reasoning. Their modular approach outperforms gAnswer and QAnswer on QALD-9 as well as QAnswer and QAMP on LC-QuAD V1. 
In contrast to seminal semantic parsing work, e.g., by Berant et al.~\cite{berant2013semantic}, we assume relatively small training data; thus, learning a generalization via semantic building blocks will not work and consequently learn the whole template on purpose. 
 
Third, there are also other QA approaches which work with neural networks relying on large, templated datasets such as sequence-to-sequence models~\cite{DBLP:journals/corr/abs-1806-10478,DBLP:journals/corr/abs-1906-09302}.
However, we do not focus on this research direction in this work. We refer the interested reader to extensive surveys and overview papers such as~\cite{DBLP:journals/corr/abs-1907-09361,DBLP:journals/kais/DiefenbachLSM18,DBLP:journals/semweb/HoffnerWMULN17}.

\section{Approach} 
\begin{figure}[htb!]
	\includegraphics[width=\linewidth]{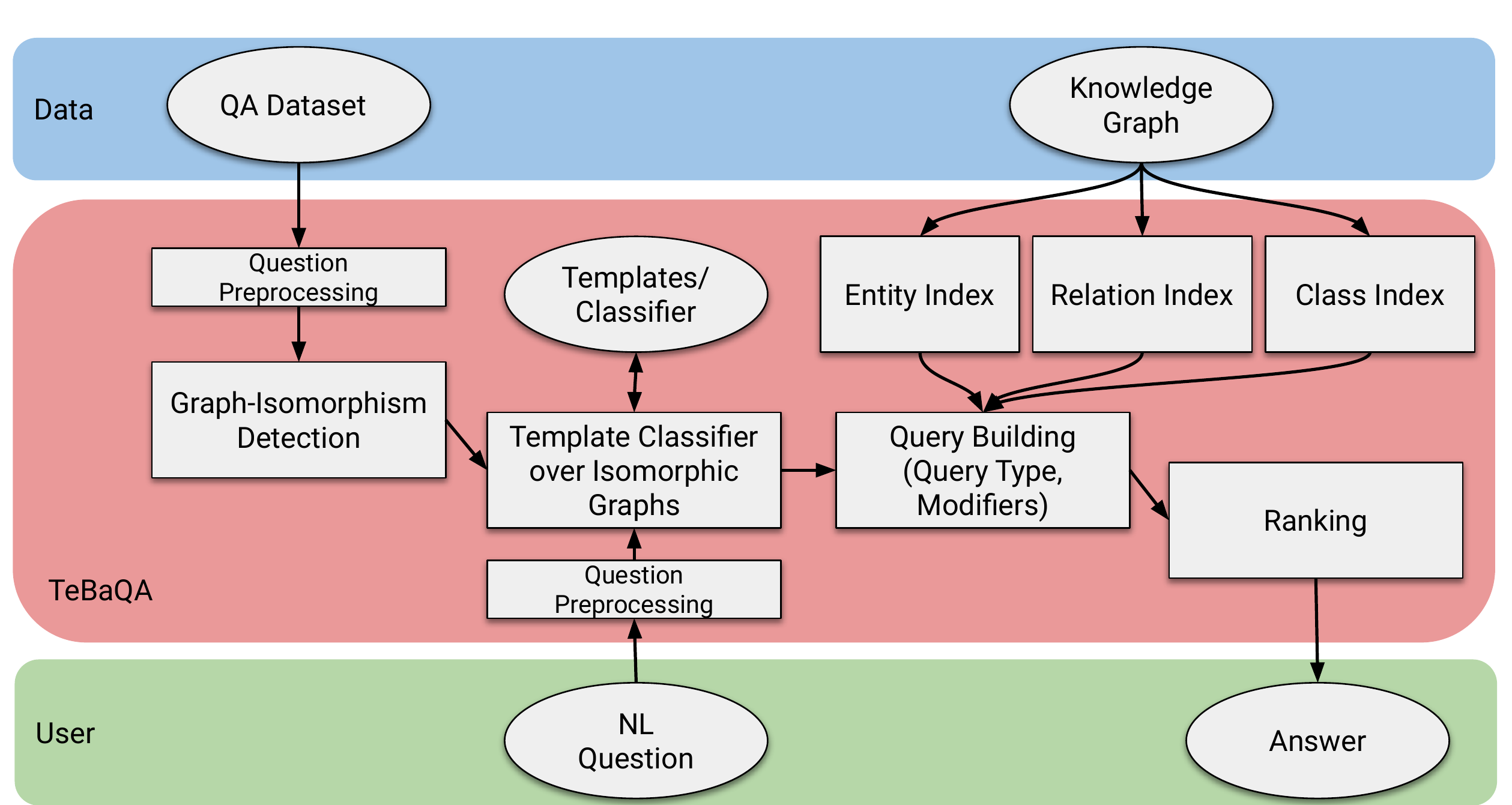}
	\caption{TeBaQA architecture on the running example.}
	\label{fig:qa-steps}
\end{figure}

TeBaQA is based on isomorphic graph patterns, which can be extracted across different SPARQL queries and hence be used as templates for our KGQA approach. Figure~\ref{fig:qa-steps} provides an overview of TeBaQA's architecture and its five main stages:

First, all questions run through a \textbf{Preprocessing} stage to remove semantically irrelevant words and create a set of meaningful n-grams.
The \textbf{Graph-Isomorphism Detection and Template Classification} phase uses the training sets to train a classifier based on a natural language question and a SPARQL query by analyzing the basic graph pattern for graph isomorphisms.
The main idea is that structurally identical SPARQL queries represent syntactically similar questions.
At runtime, a question is classified into a ranked list of SPARQL templates.
While \textbf{Information Extraction}, TeBaQA extracts all critical information such as entities, relations and classes from the question and determines the answer type based on a KG-agnostic set of indexes.
In the \textbf{Query Building} phase, the extracted information are inserted into the top templats, the SPARQL query type is determined, and query modifiers are added.
The resulting SPARQL queries are executed, and their answers are compared with the expected answer type.  The subsequent \textbf{Ranking} is based on a combination of all information, the natural language question and the returned answers.

In the following, we present each of these steps in more detail. We use DBpedia~\cite{dbpedia2014} as reference KG for the sake of simplicity in our description.

\subsection{Question Preprocessing}\label{Linguistic-Pruning}

There are often words that do not contribute any information to the answer to natural language questions.
Thus, we distinguish semantically relevant and irrelevant n-grams.
Irrelevant n-grams can lead to errors that could propagate through the architecture.
An example of this is the entity \url{dbr:The_The}\footnote{\texttt{dbr:} is a prefix which stands for \url{http://dbpedia.org/resource/}}. If the word \textit{The} were to be wrongly associated with this entity every time \textit{the} occurs in a question, the system's performance would decrease severely.
However, irrelevant words are sometimes part of entities, e.g., \url{dbr:The_Two_Towers}, so we cannot always filter these words.
For this reason, we combine up to six neighboring words from the question to n-grams and remove all n-grams that contain stop words only.
To identify irrelevant words, we provide a stop word list that contains the most common words of a particular language that are highly unlikely to add semantic value to the sentence. 
Additionally, TeBaQA distinguishes relevant and irrelevant n-grams using part-of-speech (POS) tags. Only n-grams beginning with \lstinline|JJ|, \lstinline|NN| or \lstinline|VB| POS-tags are considered as relevant. 
After this preprocessing step, TeBaQA maps the remaining n-grams to entities from DBpedia in the information extraction step.

\subsection{Graph-Isomorphism Detection and Template Classification}
TeBaQA classifies a question to determine each isomorphic basic graph pattern (BGP).
Since SPARQL is a graph-based query language~\cite{Perez:2009:SCS:1567274.1567278}, the structural equality of two SPARQL queries can be determined using an isomorphism. 
At runtime, TeBaQA can classify incoming questions to find the correct query templates, in which later semantic information can be inserted at runtime.

\subsubsection{SPARQL BGP Isomorphism to Create Template Classes:}
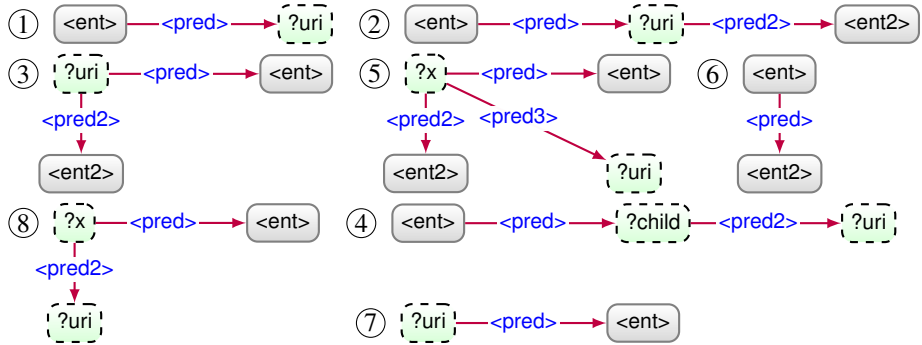
\begin{figure}[htb!]
	\begin{tabular}{l}
		\begin{tikzpicture}
    		\node[iri,anchor=center] (a) {<ent>};  
    		\node[var,anchor=center,right=2cm of a] (b) {?uri}
    		edge[arrin] node[lab] {<pred>} (a); 
    		\node[numcirc,left=0.2cm of a] {1};  
		\end{tikzpicture}\quad
		\begin{tikzpicture}
    		\node[iri,anchor=center] (a) {<ent>};
    		\node[var,anchor=center,right=2cm of a] (b) {?uri}
    		edge[arrin] node[lab] {<pred>} (a);
    		\node[iri,anchor=center,right=2cm of b] (c) {<ent2>}
    		edge[arrin] node[lab] {<pred2>} (b); 
    		\node[numcirc,left=0.2cm of a] {2};  
		\end{tikzpicture}\\
		\begin{tikzpicture}
    		\node[var,anchor=center] (a) {?uri};
    		\node[iri,anchor=center,right=2cm of a] (b) {<ent>}
    		edge[arrin] node[lab] {<pred>} (a);
    		\node[iri,anchor=center,below=0.8cm of a] (c) {<ent2>}
    		edge[arrin] node[lab] {<pred2>} (a); 
    		\node[numcirc,left=0.2cm of a] {3};  
		\end{tikzpicture}\quad
			\begin{tikzpicture}
		\node[var,anchor=center] (a) {?x};
		\node[iri,anchor=center,right=2cm of a] (b) {<ent>}
		edge[arrin] node[lab] {<pred>} (a);
		\node[iri,anchor=center,below=0.8cm of a] (c) {<ent2>}
		edge[arrin] node[lab] {<pred2>} (a); 
		\node[var,anchor=center,below=0.8cm of b] (c) {?uri}
		edge[arrin] node[lab] {<pred3>} (a); 
		\node[numcirc,left=0.2cm of a] {5};  
		\end{tikzpicture}\quad
		\begin{tikzpicture}
    		\node[iri,anchor=center] (a) {<ent>};  
    		\node[iri,anchor=center,below=0.8cm of a] (b) {<ent2>}
    		edge[arrin] node[lab] {<pred>} (a); 
    		\node[numcirc,left=0.2cm of a] {6};  
		\end{tikzpicture}\\
		\begin{tikzpicture}
        	\node[var,anchor=center] (a) {?x};
        	\node[iri,anchor=center,right=2cm of a] (b) {<ent>}
        	edge[arrin] node[lab] {<pred>} (a);
        	\node[var,anchor=center,below=0.8cm of a] (c) {?uri}
        	edge[arrin] node[lab] {<pred2>} (a); 
        	\node[numcirc,left=0.2cm of a] {8};  
	    \end{tikzpicture}\quad 
		\begin{tikzpicture}
    		\node[iri,anchor=center] (a) {<ent>};
    		\node[var,anchor=center,right=2cm of a] (b) {?child}
    		edge[arrin] node[lab] {<pred>} (a);
    		\node[var,anchor=center,right=2cm of b] (c) {?uri}
    		edge[arrin] node[lab] {<pred2>} (b); 
    		\node[numcirc,left=0.2cm of a] {4};  
    		\node[var,anchor=centerr,below=0.8cm of a] (a1) {?uri};  
    		\node[iri,anchor=center,right=2cm of a1] (b1) {<ent>}
    		edge[arrin] node[lab] {<pred>} (a1); 
    		\node[numcirc,left=0.2cm of a1] {7};  
		\end{tikzpicture}\\
		
	\end{tabular}
	\caption{All basic graph patterns used as classes for QALD-8 and QALD-9  which later become templates. Note, the depicted templates contain more than five examples in the training dataset. Our running example, "Who was the doctoral advisor of Albert Einstein?" belongs to template (1).}
	\label{fig:pattern}
\end{figure}

Using the training datasets, TeBaQA generates one basic graph pattern for each given question and its corresponding SPARQL query, see Figure~\ref{fig:pattern}.
Subsequently, all \textit{isomorphic} SPARQL queries are grouped into the same class. 
Now, each class contains semantically different natural language questions but structurally similar SPARQL queries.

\begin{theorem}[Isomorphism of labeled graphs]	\label{graph-isomorphism-theorem}
  Two labeled graphs are isomorphic when a 1:1 relationship and surjective function is present between the nodes of the graphs, wherein the node labels, edge labels, and neighborhood relationships are preserved by the mapping~\cite{gervasi2006computational}.
\end{theorem}

\subsubsection{Question Features and Classification}
Next, TeBaQA trains a classifier that uses all questions of an isomorphism class as input to calculate features for this class.  A feature vector holds all the information required to make a reliable statement about which question belongs to which class. The features can be seen in Table~\ref{table-question-features}.

\begin{table*}[htb!]
\caption{Features to map a question to an isomorphic basic graph pattern.}\label{table-question-features}
\begin{tabularx}{\textwidth}{llX}
\toprule
  \textbf{Feature} & \textbf{Type} & \textbf{Description} \\
  \midrule
  QuestionWord & Nominal & Adds the question word (e.g. Who, What, Give) as a feature. \\
  EntityPerson & Boolean & Checks the named entity tags of the sentence to see if any persons are mentioned in it.  \\
  NumberOfToken & Numeric & Stores the number of tokens separated by spaces excluding punctuation. \\
  QueryResourceType & Nominal & Categorizes the question based on a list of subject areas, e.g., film, music, book or city.\\
  Noun & Numeric & Aggregates the number of nouns. \\
  Number & Numeric & Indicates how often numbers occur in the question.\\
  Verb & Numeric & Aggregates the number of verbs. \\
  Adjective & Numeric & Aggregates the number of adjectives. \\
  Comperative & Boolean & Indicates whether comparative adjectives or adverbs are included in the sentence.\\
  TripleCandidates & Numerical & Estimates how many SPARQL triples are needed to answer the question based on the number of verbs, adjectives, and related nouns.\\
  \bottomrule
\end{tabularx}
\end{table*}

The features can be divided into semantic and syntactic features. \textit{QuestionWord}, \textit{EntityPerson} and \textit{QueryResourceType} form the group of semantic features and represent particular content aspects of the question, e.g., persons or specific topics that are mentioned in the question. All other features describe the structure of the question.

Note, other features were investigated, which did not improve the model's recognition rate. 
We report these features to aid future research in this area: 1) {Cultural Categories}: Mainly included music and movies, e.g., \textit{Who is the singer on the album The Dark Side of the Moon?} and 2) {Geographical entities}: Questions in which countries or cities occur, as well as where questions, e.g., \textit{In which country is Mecca located?}

Using features above, it is possible to represent the question \textit{Who was the doctoral advisor of Albert Einstein?} with the following vector:
\begin{lstlisting}[captionpos=b, basicstyle=\ttfamily,frame=single,basicstyle=\small]
<Who,Person,8,dbo:Person,1,0,1,0,NoComperative,1>
\end{lstlisting}

TeBaQA trains a statistical classifier using the described features extracted from the input question and the isomorphic basic graph patterns as class labels.
A feature vector's target class can be determined by generating the basic graph pattern for the corresponding SPARQL query and assigning the class, which represents this pattern.
An evaluation can be found in Section~\ref{sec:eval_classifier}.

\subsection{Information Extraction}
TeBaQA identifies entities, classes, and relations to fill the placeholders of a particular SPARQL template. 
Since questions are shorter than usual texts, semantic entity linking tools such as DBpedia Spotlight~\cite{isem2013daiber} or MAG~\cite{moussallem2017mag,MAG} do not perform correctly due to the lack of semantic context information. 
For example, in \textit{Who was the doctoral advisor of Albert Einstein?}, the word \textit{Einstein} has to be linked to \url{dbr:Albert_Einstein} and not to any other person with that name. 
For this reason, we apply a KB-agnostic search index-based approach to identify entity, relation and class candidates.  TeBaQA uses three indexes, which are created before runtime.
\subsubsection{Entity Index} The entity index contains all entities from the target knowledge graph. To map an n-gram from the preprocessing step to an entity, TeBaQA queries against the index's label field. The index contains information about entities, relations and classes connected to the entity at hand. 
\subsubsection{Relation Index and Class Index}
These two indexes contain all OWL classes and relations from a KG. The indexes are used to map n-grams to relations and classes of the ontologies of the KB.TeBaQA additionally indexes hypernyms and synonyms for all relations and classes.\footnote{The dictionary can be found at \url{https://github.com/dice-group/NLIWOD/tree/master/qa.annotation/src/main/resources} which was previously used by~\cite{DBLP:conf/esws/SinghBRS18}}

Consider the question \textit{Who was the doctoral mentor of Einstein?}. DBpedia contains only the relation \url{dbo:doctoralAdvisor}\footnote{\url{dbo:} stands for \url{http://dbpedia.org/ontology/}} and not \url{dbp:mentor}\footnote{\url{dbp:} stands for \url{http://dbpedia.org/property/}}. 
Through the synonym \textit{advisor} of \textit{mentor}, the relation \url{dbo:doctoralAdvisor} can be determined.
This example highlights the lexical and semantic gap between natural language and knowledge graphs. 

\subsubsection{Disambiguation}
By querying the indexes for an n-gram, we get candidates for entities, relations and classes, whose labels contain all tokens of the n-gram.
Since a candidate's label may contain more tokens than the n-gram, we apply a Levenshtein-distance filter of 0.8 on the candidates.
All remaining candidates are used to fill a given template.

\subsection{Query Building}
\label{query_building}
\subsubsection{Template Filling}
For filling the templates, we facilitate the information about connected entities and connected relations for the found entities from the entity-index. 
For the triples in a template, there are two cases:

1.) The triple contains one placeholder for an entity and one placeholder for a relation.
In this case, we resort to only the connected relation information from the entity index.
An entity candidate $e$ and a relation candidate $p$ are combined to a triple $<e,p,?v>$ or $<?v,p,e>$ if the set of connected relations $S(e)$ of the entity $e$ contains $p$ and if the connected n-grams do not contain each other.

2.) The triple contains only one placeholder for a relation $p'$. This case only occurs when at least one triple in the template matches case 1. We can utilize these triples to generate matching triples for the given triple.
Thus, we query the entity index and search for a set of entities $S(e')$ connected with the entity $e$ by the relation $p$. 
All connected relations from the entities in $S(e')$ in the set of relation candidates and whose n-grams do not cover the n-grams of $e$ and $p$ are candidates for $p'$.

Each candidate SPARQL query is checked for consistency with the ontology. In general, there are query patterns that do not contain variables. This case only occurs in ask queries like \textit{Did Socrates influence Aristotle?}. We ignore this case to keep simplicity and aware of the performance impact. To summarize, TeBaQA creates several candidate SPARQL queries per template and thus per question. 

\subsubsection{Query Modifiers and Query Types}
To translate a question into a semantically equivalent SPARQL query, it is often not enough to recognize the entities, relations or classes in the question and insert them into a SPARQL query. For example, the question \textit{How many children did Benjamin Franklin have?} asks for the number of children and not the concrete list.
Thus, we apply a rule-based look-up to add query modifiers and choose a query type. The supported modifiers are $COUNT$, if the question contains keywords like \textit{How many or How much}, $Filter (?x<?y)$, if we identify comparatives and $ORDER\: BY\: [ASC (?x)\:  |\: DESC (?x)]\: LIMIT \:1$, if the question contains superlatives. Additionally, we support ASK-type questions, if keywords like \textit{Is} or \textit{Are} are identified. 

Note, the templates and their respective basic graph pattern neither contain information about the query type nor information about query modifiers.
Thus, TeBaQA generates one SPARQL query per candidate SPARQL query for each cross-product of query type and modifier that is recognized.
The outcome is a list of executable queries for each input question.

\subsection{Ranking}
Since the \textit{conciseness} of an answer plays a decisive role in QA in contrast to full-text search engines, only the answer that corresponds best to the user's intention should be returned. 
Thus, all generated SPARQL queries and their corresponding answers are ranked.
This ranking is carried out in two steps. 
First, we filter by 1) the expected answer type of the question in comparison to the actual answer type of the query and by 2) the cardinality of the result set. Second, TeBaQA ranks the quality of the remaining SPARQL queries.

\subsubsection{Answer Type and Cardinality Check:}
For certain types of answer sets, only those that match the question's expected answer type are considered for the next ranking step. 
We empirically analyzed the benchmark datasets and derived a rule-based system for the most common expected answer type and their distinguishing features. 
\begin{itemize}
    \item Temporal questions usually begin with the question word \textit{When}, e.g., \textit{When was the Battle of Gettysburg?}. TeBaQA expects a date as the answer type.
    \item Decision questions mostly start with a form of \textit{Be}, \textit{Do} or \textit{Have}. The possible answer type is boolean. 
    \item Questions that begin with \textit{How much} or \textit{How many} can be answered with numbers. This also includes questions that begin with a combination of the question word \textit{How} and a subsequent adjective such as \textit{How large is the Empire State Building?}
\end{itemize}
If none of the above rules apply to a question, the result set's cardinality is checked. There are two cases:
First, when several answers are needed to answer a question fully. Consider \textit{Which ingredients do I need for carrot cake?}, if only one answer is found for this question, it can be assumed that it is either wrong or incomplete. 
Second, when there is only one answer to a question, e.g., \textit{In which UK city are the headquarters of the MI6?}, an answer consisting of several entities would not be correct.

To recognize to which query type (ASK or SELECT) a question belongs, the first noun or the first compound noun after the question word is checked. 
If they occur in the singular form, a single answer is needed to answer the question. For the above question \textit{In which UK city are the headquarters of the MI6?}, the compound noun would be \textit{UK city}. Since both words occur in the singular, it is assumed that only a single answer is required. 
If the first noun or group of nouns occurs in the plural, this indicates that the question requires multiple answers. In the question \textit{Which ingredients do I need for carrot cake?} the decisive word \textit{ingredients} is in the plural.

However, the answer type of question may not be recognized correctly. For instance, if the question is grammatically correct but words whose singular form is identical to the plural form exist, such as \textit{news}, we cannot determine the correct answer type. These issues will be tackled in future research.

Once the type of question and answer has been determined, all the answers whose type or cardinality does not match the question will be discarded.

\subsubsection{Quality Ranking:}
For the remaining SPARQL queries, TeBaQA calculates a $rating$ based on the sum of the individual scores of the bindings $B$ and the input question $phrase$. A binding $B$ is the mapping of entities, relations and classes to placeholders contained in one SPARQL query $q$.  To compute the {relatedness factor} $r$, the following factors are taken into account:
\begin{itemize}
	\item \textbf{Annotation Density}: The annotation density measures that the more words from the sentence are linked to an entity, class or relation, the more likely it is that it corresponds to the intention of the user.  For the question \textit{What is the alma mater of the chancellor of Germany Angela Merkel?} one candidate query may apply the binding \url{dbr:Angela}, while another query applies the binding \url{dbr:Angela_Merkel}. The former refers only to the word \textit{Angela}. The latter refers to two words of the sentence: \textit{Angela Merkel} and covers longer parts of the $phrase$.
	
	\item \textbf{Syntactic Similarity}: The syntactic similarity is an indicator of how similar a n-gram of the sentence and the associated binding are. For example, in the question \textit{Who is the author of the interpretation of dreams?} the n-gram \textit{the interpretation of dreams} can be linked with \url{dbr:The_Interpretation_of_Dreams} or \url{dbr:Great_Book_of_Interpretation_of_Dreams} among others. The former has a smaller Levenshtein distance and a greater syntactic similarity with the selected n-gram.
\end{itemize}

We cover both aspects with the following formulas:
\[ rating = \sum_{B \in q} r(B, phrase)\]

\[r(entity, phrase) = |words(phrase)| - levenshtein\_ratio(label(B), phrase)\]

After all entities, classes, and relations used in a query have been evaluated and summed up, the rating is corrected down by 30\% if more than 50 results are returned by the query, based on empirical observations in the datasets.

\section{Evaluation}\label{chapter:evaluation}
\subsection{Datasets}

We performed the evaluation on the 8th and 9th Question Answering over Linked Data challenge training datasets (QALD-8 train~\cite{qald-8} and  QALD-9 train~\cite{qald-9}), which contain 220 (QALD-8) and 408 (QALD-9) heterogeneous training questions. Additionally, we evaluated on the two LC-QuAD~\cite{LC-QuADv1,lc-quadv2} datasets with 4000 train and 1000 test questions and 24.000 train and 6.000 test questions, respectively. Across datasets, the questions vary in complexity since they also include comparatives, superlatives, and temporal aggregations. An example of a simple question is \textit{How tall is Amazon Eve?}. A more complex example is \textit{How many companies were founded in the same year as Google?} since it contains a temporal aggregation (\textit{the same year}).
We created separate instances of TeBaQA for each of the training datasets and evaluated each instance on the corresponding test dataset.

\subsection{Classification Evaluation}
\label{sec:eval_classifier}

For the QALD-8 and QALD-9 datasets, eight classes were identified for each dataset, compare Figure~\ref{fig:pattern}.
For LC-QuAD v1 and LC-QuAD v2, TeBaQA identified 17 classes for both datasets. Since the two LC-QuAD datasets were constructed for more diversity, the classification is more challenging.
Note, we omitted classes with less than five examples in the training dataset. We are aware that we are trading our overall performance for classification accuracy. 

To this end, we evaluated a variety of machine learning methods, which required questions to be converted into feature vectors.
In particular, we used the QALD-8 and QALD-9 training datasets as our training data and used 10-fold cross-validation to evaluate the computed models. 
All duplicate questions and questions without SPARQL queries were removed from the training datasets.
We tested multiple machine learning algorithms with the WEKA framework\footnote{\url{https://www.cs.waikato.ac.nz/ml/weka/}}~\cite{eibe2016weka} on our training data using 10-fold cross-validation.
To achieve comparable results, TeBaQA uses only the standard configuration of the algorithms.
The macro-weighted F-Measure for one fold in cross-validation is calculated from the classes' F-Measures, weighted according to the size of the class. 
After that, we calculated the average macro-weighted F-Measure across all folds. 
On the QALD datasets, the algorithm \textit{RandomizableFilteredClassifier} achieves the highest F-Measures of 0.523964 (QALD-8) and 0.528875 (QALD-9), respectively. 
On LC-QuAD v1, we achieve a template classification f-measure of 0.400464 and 0.425953 on LC-QuAD v2
using a MultilayerPerceptron. 
Consequently, we use the best performing classifiers for the end-to-end evaluation.

Similar experiments can be found in Athreya et al.'s work~\cite{athreya2020templatebased}. 
The authors use a recurrent neural network, i.e., a tree-LSTM, to identify templates in LC-QuAD and achieve an accuracy of 0.828 after merging several template classes manually.

\subsection{GERBIL QA Benchmark}\label{sec:gerbil-benchmark}

For evaluation, we used the FAIR benchmarking platform GERBIL QA~\cite{DBLP:journals/semweb/UsbeckRHCHNDU19} to ensure future reproducibility of the experiments.

Table~\ref{table:qa-systeme-vergleich} contains the results of selected Question Answering systems, measured against the QALD-8 and the QALD-9 test benchmarks\footnote{The links to our GERBIL-experiments can be found on our Github page: \url{https://github.com/dice-group/TeBaQA/blob/master/README.md}}. We focused on English questions only, as English is supported by all available QA systems at the time of these experiments. The macro value for Precision, Recall and F-Measure was selected. The evaluation was performed with GERBIL version 0.2.3 if possible. We report always the highest numbers if several papers reported numbers and evaluation with GERBIL was not possible. 
\begin{table}[htb!]
    \setlength{\tabcolsep}{6pt}
    	\centering
    	\caption{Results of TeBaQA and other state-of-the-art QA systems for multiple datasets. * indicates F-1 measure instead of QALD F-Measure~\cite{DBLP:journals/semweb/UsbeckRHCHNDU19}. ** Numbers taken from the corresponding paper.}
    	\label{table:qa-systeme-vergleich}
        \resizebox{\textwidth}{!}{%
        	\begin{tabular}{lccrrrr} 
        		\toprule
        		\textbf{System}&
        		\textbf{KB}&
        		\textbf{Dataset}& 
        		\textbf{Precision} & \textbf{Recall} &  \textbf{QALD  F-Measure} & \textbf{Avg. Time in s}\\
        	\midrule
        		gAnswer2~\cite{DBLP:conf/sigmod/ZouHWYHZ14} & DBpedia& QALD-8    & 0.337 & 0.354 & 0.440 &	4.548  \\ 
        		QAnswer~\cite{DBLP:journals/semweb/DiefenbachGBSM20} & DBpedia& QALD-8     & 0.452 & 0.480 & 0.512 & 0.446   \\
        		Zheng et al.~\cite{zheng2019question}**     & DBpedia& QALD-8 &0.459  & 0.463& * 0.461 &- \\
        	    TeBaQA & DBpedia& QALD-8 & 0.476 &	0.488 & \textbf{0.556} & 28.990  \\
    		\midrule
        		Elon~\cite{qald-9}  & DBpedia   & QALD-9& 0.049 & 0.053 & 0.100 & 0.219   \\ 
        		gAnswer~\cite{DBLP:conf/sigmod/ZouHWYHZ14} & DBpedia & QALD-9 & 0.293 & 0.327 & 0.430 & 3.076    \\ 
        		NSQA~\cite{DBLP:journals/corr/abs-2012-01707}** & DBpedia& QALD-9 & 0.314 & 0.322 & *0.453 & - \\
        		QAnswer~\cite{DBLP:journals/semweb/DiefenbachGBSM20} & DBpedia & QALD-9& 0.261 & 0.267 & 0.289 & 0.661 \\
        		QASystem~\cite{qald-9} & DBpedia & QALD-9& 0.097 & 0.116 & 0.200 & 1.014  \\ 
        		Zheng et al.~\cite{zheng2019question}**     & DBpedia& QALD-9 &0.458  & 0.471 & \textbf{*0.463} &- \\
        	    TeBaQA   & DBpedia& QALD-9 & 0.241 & 0.245 & 0.374 & 5.049  \\
            \midrule
        		NSQA~\cite{DBLP:journals/corr/abs-2012-01707}**       & DBpedia& LC-QuAD v1& 0.382  &  0.404 &   *0.383 & - \\
        		QAMP~\cite{DBLP:conf/cikm/VakulenkoGPRC19}**          & DBpedia& LC-QuAD v1 & 0.250 &  0.500  &  *0.330 & 0.720   \\  
        		QAnswer~\cite{DBLP:journals/semweb/DiefenbachGBSM20} **      & DBpedia& LC-QuAD v1 &  0.590  & 0.380  & \textbf{*0.460}   & 1.500
        		\\
                TeBaQA                                              & DBpedia& LC-QuAD v1&0.230 & 0.229 & 0.300 & 36.000 \\
             \midrule
             TeBaQA  & Wikidata& LC-QuAD v2 & 0.140 & 0.136 & 0.227  & -- \\
    		\bottomrule
        	\end{tabular}
        }
\end{table}

On the QALD-8 benchmark, TeBaQA achieved the best results in terms of F-Measure by 5\% QALD F-measure. Our average time is significantly larger than the other reported systems since the ranking mechanism gets activated and then fires several SPARQL queries after the initial null-retrieving SPARQL query. 

On QALD-9, TeBaQA is in fourth place with a QALD F-Measure of 0.37. This implies that TeBaQA achieves comparable or partially better results than other semantic QA systems with a wide margin of possible improvements, as shown in the ablation study.
QALD-9 is a more challenging benchmark than QALD-8 since it contains many questions that require complex queries with more than two triples. 
A more in-depth analysis shows that questions from QALD-9 often require complex templates that are not contained in the training queries or have only low support in the training set~\cite{DBLP:journals/corr/abs-2011-07743}. This mismatch leads to a high number of misclassifications and explains the limited performance on QALD-9 compared to QALD-8 and, thus, its limited generalization abilities to unseen templates.
Although we are not outperforming the state-of-the-art systems on QALD-9, TeBaQA a novel research avenue w.r.t. learning from data.

On the LC-QuAD dataset, which contains the most complex questions, TeBaQA achieves an F-Measure of 0.30. We ran this benchmark only once in our system and had some errors during runtime. We will further investigate the performance on LC-QuAD in our future research.
Note, we ran LC-QuAD only once through the system to test our independence of the dataset similar to the methodology of Berant et al.~\cite{wang-etal-2015-building}.

To the best of our knowledge and after reaching out to the authors of LC-QuAD, we are the first system to evaluate on LC-QuAD v2. Since LC-Quad v2 uses Wikidata instead of DBpedia, we generated separate indexes for the Wikdata KG. The rest of the system remained unchanged. This shows, that TeBaQA can be easily adapted to other Knowledge Graphs.

\subsection{SPARQL operation modifiers}

The Achilles heel of most QA systems is their inability to deal with SPARQL operation modifiers, e.g., questions involving aggregations or superlatives~\cite{DBLP:conf/semweb/0002DUN17}.
In contrast to those other systems, TeBaQA contains functionalities to identify the query type and other modifiers. 
We analyzed the results of TeBaQA on these benchmark questions and found that TeBaQA was able to answer many of those questions, while other systems like QAswer and gAnswer fail on this; see supplementary material. 

\subsection{Ablation Study} \label{ablation}

\begin{table}[htb!]
    \setlength{\tabcolsep}{6pt}
	\centering
	\caption{Ablation study for TeBaQA on the QALD-9 test benchmark.}
	\label{table:ablation_study}
	 \resizebox{\textwidth}{!}{%
	\begin{tabular}{lrrrr} 
		\toprule
		\textbf{QA System}             & \textbf{Precision} & \textbf{Recall} &  \textbf{Avg. Time} & \textbf{QALD F-Measure} \\ 
		\midrule
		Perfect classification & 0.205 & 0.210 & 4.618 & 0.337  \\
		Perfect classification + EL & 0.407 & 0.407 & 0.355 & 0.578 \\ 
		Perfect classification + ranking & 0.245 & 0.257 & 4.029 & 0.399  \\ 
		Perfect EL & 0.301 & 0.317 & 0.713 & 0.473\\ 
		Perfect EL + ranking &  0.302 & 0.320 & 0.653 & 0.477 \\ 
		Perfect ranking & 0.258 & 0.270 & 6.281 & 0.405 \\ 
		Perfect classification + EL + ranking & 0.407 & 0.407 & 0.251 & 0.578 \\ 
		\bottomrule
	\end{tabular}
	}
\end{table}

We performed an ablation study to find the modules of TeBaQA's pipeline, which influence the end-to-end performance the most. Since the number of possible entities is roughly a magnitude larger than the number of relations and for the sake of experiment time, we omitted to test for perfect relation and class linking. 

For the perfect classification experiment, the QALD F-Measure is lower than for the overall system, see Table~\ref{table:qa-systeme-vergleich}.
Investigating the detailed outputs, TeBaQA selects the simple templates containing only one triple more often than the more complex templates because they have more instances, i.e., support, on the QALD-9 train dataset. In many cases, a simple query generates a result set that is a superset of the target result set and, in consequence, decreases the precision. 

When TeBaQA fails to fill the correct complex template with the correct entities, the query result set is often disjoint from the target result set. 
It is also reasonable that TeBaQA fails to fill the more complex templates due to missing or incorrect entity links.

Still, there is a gap in the system, which becomes evident when looking at the perfect classification plus ranking. Ranking gets only activated if the perfect template filled with semantic information does not retrieve any answer. That is, TeBaQA fails to find the correct modifiers or needs to circle through other semantic information candidates.

When the perfect classification is combined with perfect entity linking, the results reach a QALD F-measure of 0.58, which clearly would outperform any other system. The same happens if we add the perfect ranking. The results are the same in both cases because the ranking is most often not triggered since the perfect template is already filled correctly. 

The strongest, single influence is the entity linking part, enabling TeBaQA to jump to 0.47 F-measure. We will tackle this challenging module in future work.

Regarding runtime, failing to find the perfect template and then iterating through the ranking, i.e., querying the SPARQL endpoint often, increases the average time needed significantly.

\section{Summary and Future Work}

We presented TeBaQA, a QA system which learns to map question to SPARQL template mappings using basic graph pattern isomorphisms. 
TeBaQA significantly eases the domain/KB adoption process as it relies only on a benchmark dataset at its core. 
 
In the future, we will evaluate TeBaQA on more heterogeneous KG benchmarks to identify further choke points. We will also improve the question classification and information extraction by using novel deep learning mechanism. 

\textbf{Acknowledgements} We acknowledge the support of the Federal Ministry for Economic Affairs and Energy (BMWi) project SPEAKER (FKZ 01MK20011A), ScaDS.AI (01/S18026A) as well as the Fraunhofer Zukunftsstiftung project JOSEPH and the Eurostars Project PORQUE (FKZ 01QE2056C). This work was partially supported by the German Federal Ministry of Transport and Digital Infrastructure (BMVI) in the project LIMBO (no. 19F2029I) and by the German Federal Ministry of Education and Research (BMBF) in the project SOLIDE (no. 13N14456) within ’KMU-innovativ: Forschung f\"ur die zivile Sicherheit’ in particular ’Forschung f\"ur die zivile Sicherheit’.

\bibliographystyle{abbrv}
\bibliography{bibliography.bib}

\end{document}